\documentclass{article}

\usepackage{amsmath,amssymb,amsfonts}
\usepackage{algorithm}
\usepackage{algorithmic}
\usepackage{graphicx}
\usepackage{booktabs}
\usepackage{hyperref}
\usepackage{url}
\usepackage{xcolor}
\usepackage{microtype}
\usepackage{natbib}

\newcommand{\Dt}{\mathcal{D}_t}
\newcommand{\Ht}{\mathcal{H}}
\newcommand{\piLLM}{\pi_{\mathrm{LLM}}}
\newcommand{\mut}{\mu_t}
\newcommand{\sigt}{\sigma_t}

\usepackage{authblk}

\title{BayesEvolve: Explicit Belief States for Autonomous Scientific Discovery}

\author{
Xuening Wu,$^{1}$ 
Shan Yu,$^{2}$ 
Qianya Xu,$^{3}$ 
Shenqin Yin$^{4,†}$\\
{\small
$^{1}$Pfizer, Shanghai, China\\
$^{2}$Independent Researcher, Hangzhou, China\\
$^{3}$University of California San Diego, La Jolla, CA, USA\\
$^{4}$Institute of Humanities and Social Science Data, Fudan University, Shanghai, China\\
$^{†}$Corresponding author: \texttt{ysq@fudan.edu.cn}}}

\date{} 

\begin{document}
\maketitle
\begin{abstract}
Autonomous scientific discovery systems increasingly use large language models (LLMs) to propose new hypotheses, but many such systems condition primarily on experimental memory: archives of high-scoring candidates or heuristic summaries of recent trials. We argue that discovery agents should instead maintain explicit, uncertainty-aware beliefs about hypothesis quality. We introduce \textbf{BayesEvolve}, a belief-guided discovery framework that converts experimental evidence into a predictive belief state and uses this belief to guide future experimentation. As a controlled testbed for belief-guided discovery, we evaluate BayesEvolve on shifted BBOB-style black-box optimization tasks, leaving program and laboratory discovery domains to future work. BayesEvolve improves sample efficiency over memory- and archive-guided LLM baselines under a fixed evaluation budget. We further show that the belief state is predictive on held-out candidate pools, that controlled decision-rule ablations favor belief-guided selection with an annealed uncertainty bonus, and that BayesEvolve exhibits productive late-stage concentration rather than unfocused exploration.
\end{abstract}

\section{Introduction}
\label{sec:intro}

LLM-guided evolutionary search has emerged as a compelling paradigm for automated optimization and discovery. FunSearch~\citep{romera2024funsearch} demonstrates that evolutionary selection over LLM-generated programs can recover novel combinatorial results; AlphaEvolve~\citep{novikov2025alphaevolve} scales similar ideas to larger codebases and optimization tasks. Despite these successes, many systems use experimental history primarily as memory: the LLM is prompted with examples sampled from an archive of previously evaluated candidates, often biased toward high-scoring examples.

This archive-guided design leaves much of the evidence in implicit form. A list of past experiments says what happened, but it does not explicitly encode what the agent currently believes about unevaluated hypotheses, how uncertain those beliefs are, or which experiment would be most informative next. Two costs follow. First, the agent cannot directly reason about predicted quality beyond the observed archive. Second, it may either over-concentrate around early elite candidates or continue exploring without converting evidence into a focused search direction.

We propose \textbf{BayesEvolve}, which reframes autonomous discovery as belief-state evolution. Instead of treating past evaluations as unstructured memory, BayesEvolve maintains an explicit predictive belief state over candidate quality. At each step, experimental evidence updates a posterior belief; this belief is exposed to the proposal process and used by an acquisition rule to select future evaluations. In the experiments below, the belief state is implemented with a Gaussian process (GP) posterior over numerical candidates, but the framework is agnostic to the surrogate model and candidate representation.

\paragraph{Contributions.}
\begin{itemize}
  \item We formulate LLM-guided discovery as explicit belief-state maintenance rather than archive memory alone.
  \item We introduce a belief-guided selection rule with an annealed uncertainty bonus that shifts from exploration to exploitation as evidence accumulates.
  \item We evaluate BayesEvolve on shifted BBOB-style optimization tasks, showing stronger performance than archive- and memory-guided LLM baselines.
  \item We analyze belief quality, decision-rule ablations, and diversity dynamics, showing that BayesEvolve's belief state is predictive and supports productive late-stage concentration.
\end{itemize}

\section{Related Work}
\label{sec:related}

\paragraph{Archive-guided LLM evolutionary search.}
FunSearch~\citep{romera2024funsearch} evolves programs using a population of evaluated candidates, while AlphaEvolve~\citep{novikov2025alphaevolve} extends LLM-guided evolution to broader engineering problems. These systems demonstrate the power of LLM proposal plus selection, but their historical information is largely represented through archives and scores. BayesEvolve complements this line of work by making predictive beliefs explicit.

\paragraph{Bayesian and probabilistic discovery.}
Bayesian optimization uses posterior beliefs and acquisition functions to select informative evaluations~\citep{jones1998efficient,srinivas2010gaussian}. BayesEvolve borrows this decision-theoretic structure but places it inside an LLM-guided discovery loop, where the belief state is used both to summarize evidence and guide candidate selection. Related probabilistic discovery systems such as ModelSMC~\citep{wahl2026modelsmc} maintain distributions over symbolic model candidates; BayesEvolve focuses on explicit predictive beliefs for general hypothesis quality.

\paragraph{Memory-guided LLM agents.}
Memory buffers and archive summaries are common in LLM-based scientific agents and optimization systems. Such memory can be useful, but it does not by itself provide calibrated predictions or uncertainty estimates over unevaluated candidates. Our experiments compare against archive and heuristic-memory baselines to isolate the value of explicit belief states.

\section{BayesEvolve}
\label{sec:method}

\subsection{Problem Formulation}
\label{sec:formulation}

Let $\Ht$ be a hypothesis space. At step $t$, the system evaluates a candidate $h_t \in \Ht$ and receives an objective value $y_t \in \mathbb{R}$. In this paper we consider minimization, so lower $y_t$ is better. The evaluation history is
\begin{equation}
  \Dt = \{(h_i, y_i)\}_{i=1}^{t},
\end{equation}
and the goal is to find a candidate with low objective value under a fixed evaluation budget $T$.

Archive-guided methods condition proposals on a subset of past evaluations, such as top-scoring candidates or recent experimental summaries. BayesEvolve instead maintains an explicit belief state
\begin{equation}
  P(y \mid h, \Dt) = \mathcal{N}(\mut(h), \sigt^2(h)),
\end{equation}
where $\mut(h)$ is the predicted objective and $\sigt(h)$ is posterior uncertainty.

\subsection{Belief-Guided Selection}
\label{sec:acq}

For a candidate pool $\mathcal{C}_t$, BayesEvolve selects the next candidate using an uncertainty-aware score. Because we minimize the objective, the fixed-UCB rule is
\begin{equation}
  a_t(h) = -\mut(h) + \beta\,\sigt(h).
\end{equation}
Motivated by exploration schedules in reinforcement learning, our final decision rule uses a decaying uncertainty coefficient:
\begin{equation}
  a_t(h) = -\mut(h) + \beta_t\,\sigt(h),
  \qquad
  \beta_t = \beta_0 \sqrt{\frac{n_0}{t}},
\end{equation}
where $n_0$ is the number of shared initialization evaluations. This encourages broader exploration early and increasingly exploits the belief mean as evidence accumulates.

\begin{algorithm}[t]
\caption{BayesEvolve}
\label{alg:bayesevolve}
\begin{algorithmic}[1]
\REQUIRE Proposal model $\piLLM$, surrogate model, budget $T$, initialization size $n_0$
\STATE Evaluate $n_0$ initial candidates to form $\Dt$
\FOR{$t=n_0+1$ \textbf{to} $T$}
  \STATE Fit/update belief state $P(y \mid h, \Dt)$
  \STATE Construct candidate pool $\mathcal{C}_t$ from LLM proposals and/or archive mutations
  \STATE Compute $\mut(h)$ and $\sigt(h)$ for $h \in \mathcal{C}_t$
  \STATE Select $h_t = \arg\max_{h\in\mathcal{C}_t}[-\mut(h)+\beta_t\sigt(h)]$
  \STATE Evaluate $y_t = f(h_t)$ and update $\Dt \leftarrow \Dt \cup \{(h_t,y_t)\}$
\ENDFOR
\RETURN best candidate in $\Dt$
\end{algorithmic}
\end{algorithm}

\section{Experimental Evaluation}
\label{sec:exp}

\subsection{Setup}

\paragraph{Benchmark.}
We evaluate on five shifted BBOB-style minimization functions~\citep{hansen2009bbob} in dimension $d=5$: Sphere, Ellipsoid, Rastrigin, Rosenbrock, and Ackley. Each function is shifted by a fixed hidden offset, and LLM prompts expose only opaque task IDs (e.g., F01) rather than function names, preventing benchmark-name leakage. Each run uses $n_0=6$ shared random initialization evaluations and a total budget of $T=100$ evaluations. Results are averaged over five random seeds and five functions; shaded regions and $\pm$ values report standard error.

\paragraph{Methods.}
All LLM-based methods use the same proposal model, \texttt{gpt-5.4-mini}, and differ only in the information provided in context: no memory (Random-LLM), top archive entries (Archive-LLM), recent heuristic memory (Memory-LLM), or BayesEvolve's explicit belief state. GP-BO is a non-LLM Bayesian optimization baseline. For local ablations, all belief variants use the same shifted benchmark, shared initialization, GP posterior, and candidate-pool mechanism; we compare mean-only selection, UCB~\citep{srinivas2010gaussian}, Thompson sampling~\citep{thompson1933likelihood}, and expected improvement~\citep{jones1998efficient}.

\subsection{Experiment 1: Main Discovery Performance}

\begin{figure}[t]
\centering
\includegraphics[width=0.78\linewidth]{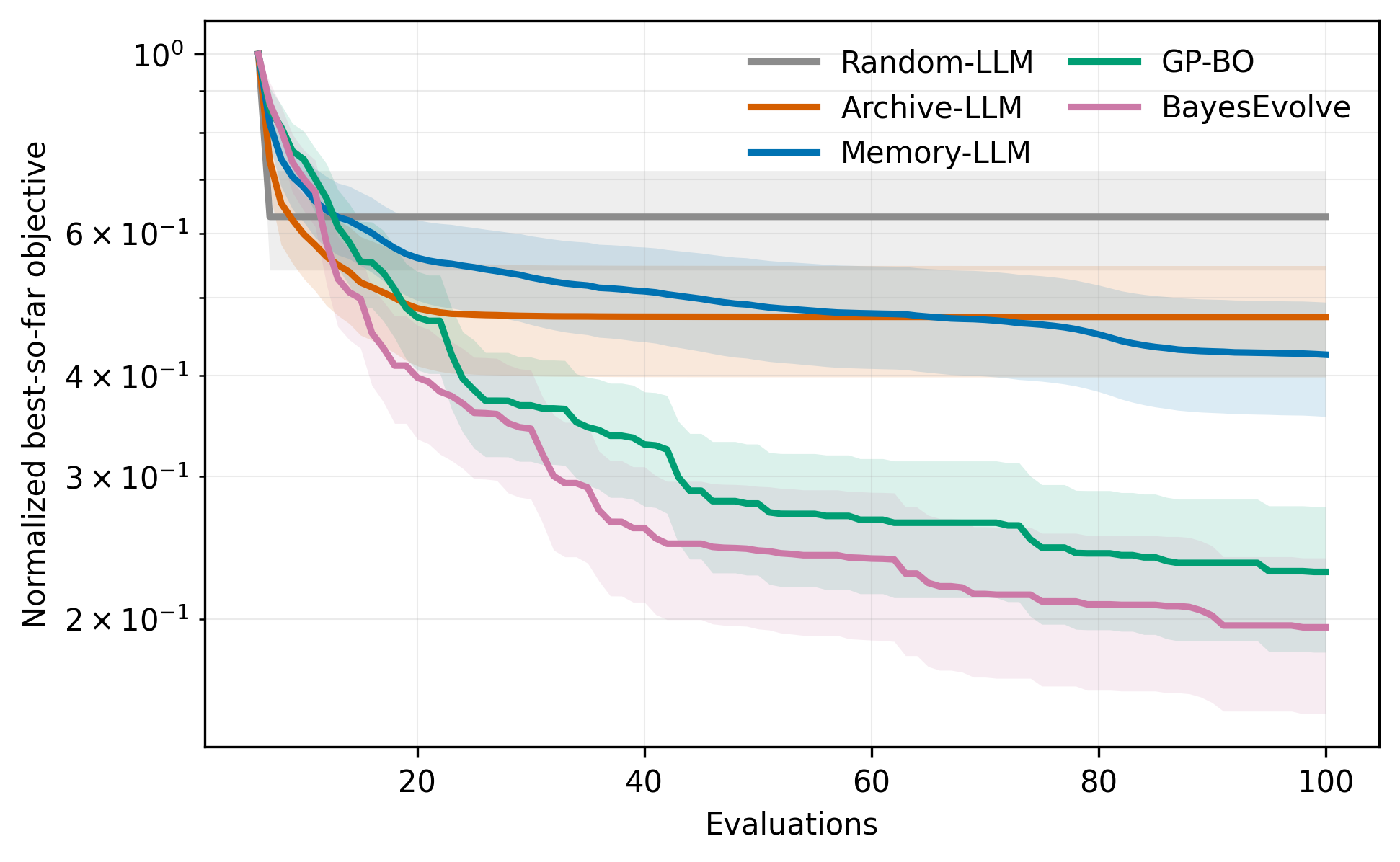}
\caption{Main discovery performance on shifted BBOB-style optimization tasks. Curves show mean normalized best-so-far objective across five benchmark functions and five random seeds; shaded regions denote standard error. All LLM methods use \texttt{gpt-5.4-mini}; GP-BO is a non-LLM Bayesian optimization baseline. Lower is better.}
\label{fig:main_perf}
\end{figure}

\begin{table}[t]
\centering
\caption{Normalized best-so-far objective at 25, 50, and 100 evaluations. Lower is better.}
\label{tab:main_perf}
\begin{tabular}{lccc}
\toprule
Method & 25 evals & 50 evals & 100 evals \\
\midrule
Random-LLM & $0.629 \pm 0.089$ & $0.629 \pm 0.089$ & $0.629 \pm 0.089$ \\
Archive-LLM & $0.476 \pm 0.074$ & $0.473 \pm 0.074$ & $0.473 \pm 0.074$ \\
Memory-LLM & $0.545 \pm 0.065$ & $0.488 \pm 0.069$ & $0.425 \pm 0.069$ \\
GP-BO & $0.384 \pm 0.059$ & $0.278 \pm 0.051$ & $0.229 \pm 0.047$ \\
\textbf{BayesEvolve} & $\mathbf{0.360 \pm 0.062}$ & $\mathbf{0.243 \pm 0.049}$ & $\mathbf{0.195 \pm 0.043}$ \\
\bottomrule
\end{tabular}
\end{table}

BayesEvolve achieves the best mean normalized objective throughout the budget and the best final performance at 100 evaluations. The gains are largest relative to archive- and memory-guided LLM baselines, suggesting that explicit predictive beliefs provide more useful guidance than experimental memory alone.

\subsection{Experiment 2: Belief State Quality}

\begin{figure}[t]
\centering
\includegraphics[width=0.78\linewidth]{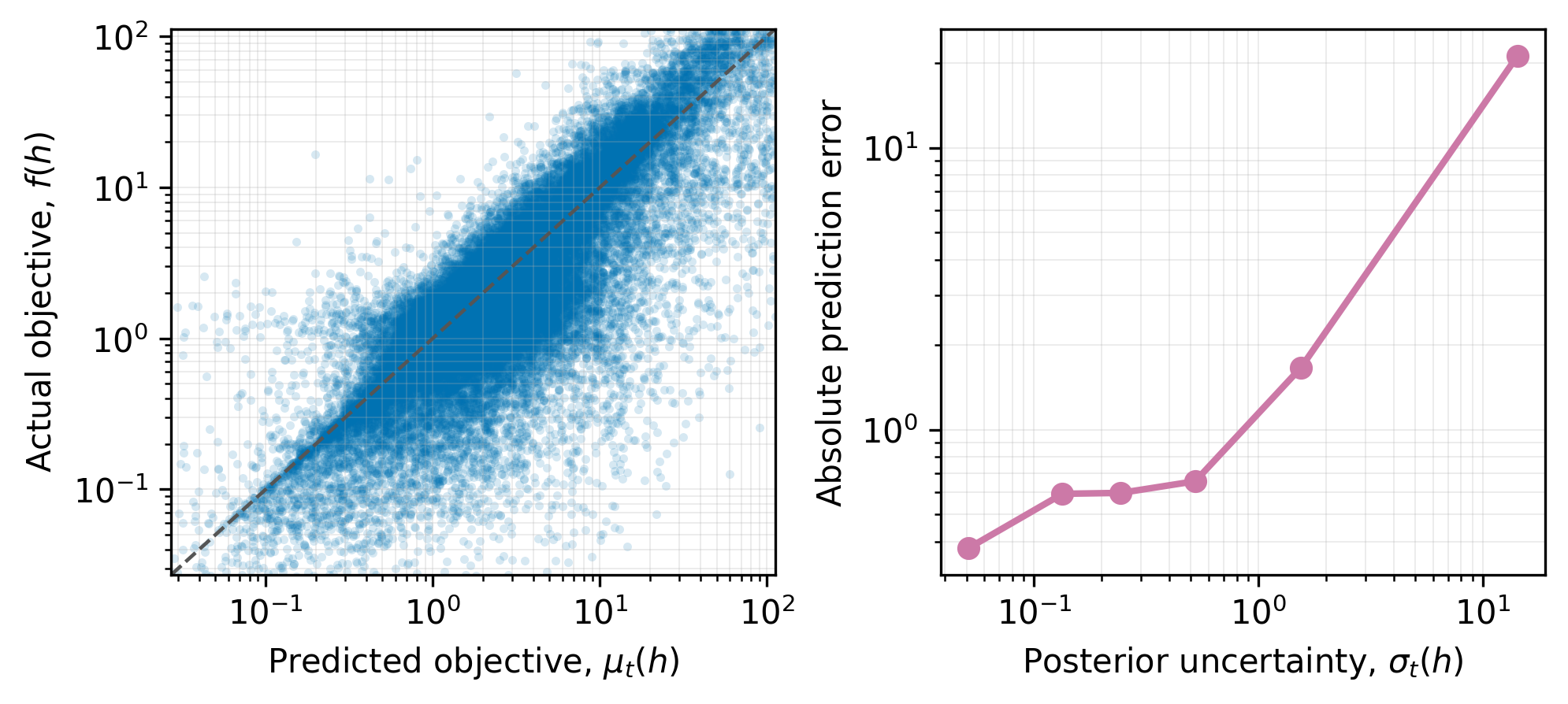}
\caption{Belief-state quality. BayesEvolve's explicit belief state is evaluated on held-out candidate pools during shifted BBOB-style optimization. The left panel compares posterior mean predictions with realized objective values; the right panel shows that candidates with higher posterior uncertainty have larger prediction error. Lower prediction error is better.}
\label{fig:belief_quality}
\end{figure}

Across 60{,}800 held-out belief predictions, posterior mean is strongly rank-correlated with realized objective values (Spearman $\rho=0.774$). Uncertainty is also informative: bins with larger posterior $\sigma_t(h)$ exhibit larger absolute prediction error. The belief state is predictive but not perfectly calibrated, with empirical coverage of $0.433$ for one-sigma intervals and $0.653$ for two-sigma intervals.

\subsection{Experiment 3: Decision Rule Ablation}

\begin{table}[t]
\centering
\caption{Decision-rule ablation. All variants use the same shifted benchmark and shared initialization. Final normalized best is lower-is-better; AUC improvement is higher-is-better.}
\label{tab:decision_ablation}
\begin{tabular}{lccc}
\toprule
Variant & Final normalized best $\downarrow$ & AUC improvement $\uparrow$ & Evals to target $\downarrow$ \\
\midrule
Archive & $0.333 \pm 0.059$ & $0.508 \pm 0.056$ & 38.6 \\
Mean-only & $0.177 \pm 0.047$ & $\mathbf{0.746 \pm 0.048}$ & \textbf{16.3} \\
Fixed UCB & $0.176 \pm 0.050$ & $0.731 \pm 0.050$ & 21.2 \\
\textbf{UCB-decay / BayesEvolve} & $\mathbf{0.174 \pm 0.045}$ & $0.730 \pm 0.046$ & 19.9 \\
Thompson & $0.210 \pm 0.048$ & $0.664 \pm 0.045$ & 24.6 \\
EI & $0.205 \pm 0.048$ & $0.672 \pm 0.054$ & 16.5 \\
GP-BO & $0.213 \pm 0.043$ & $0.669 \pm 0.049$ & 22.9 \\
\bottomrule
\end{tabular}
\end{table}

The ablation shows that explicit predictive beliefs drive most of the gain: Mean-only selection already substantially improves over archive search. Adding an annealed uncertainty bonus yields the best final score, suggesting that uncertainty is most useful when exploration is gradually reduced over the discovery process.

\subsection{Experiment 4: Diversity Dynamics}

\begin{figure}[t]
\centering
\includegraphics[width=0.95\linewidth]{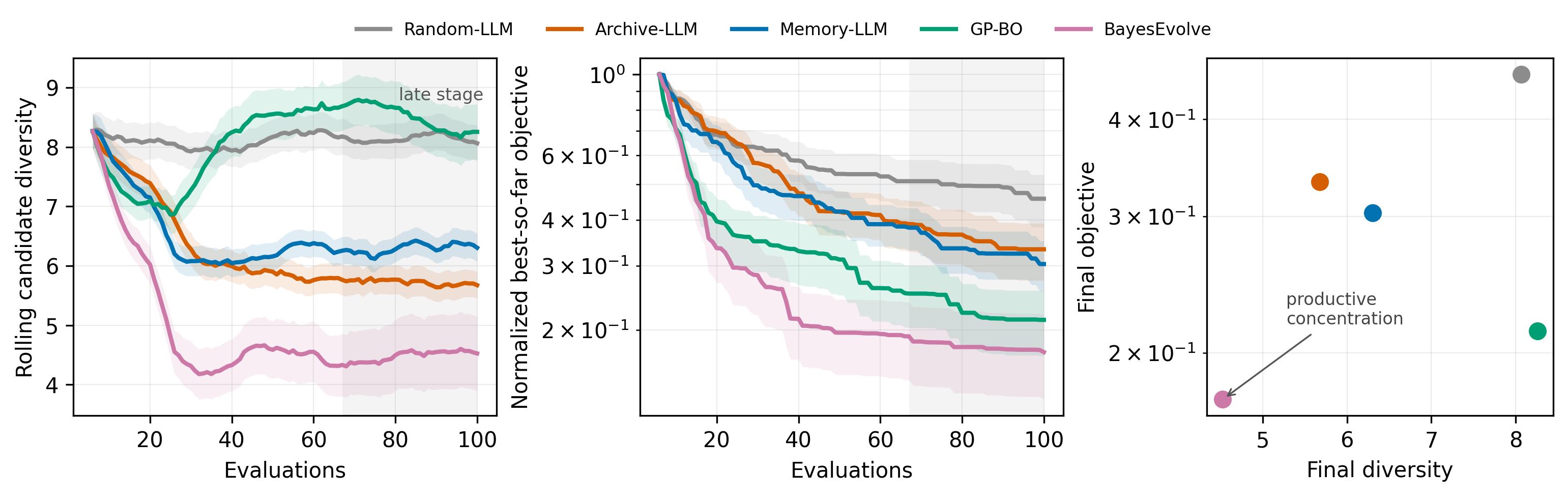}
\caption{Diversity dynamics and productive concentration. Rolling candidate diversity is computed over the most recent 20 candidates. High diversity alone is not sufficient: Random-LLM and GP-BO maintain broad exploration but do not achieve the best final objective. BayesEvolve reduces diversity in the late stage while achieving the lowest objective, indicating productive belief-guided concentration rather than unfocused exploration. Lower objective is better.}
\label{fig:diversity}
\end{figure}

Diversity is not intrinsically better when maximized throughout the run. Random-LLM and GP-BO maintain high final diversity, but BayesEvolve achieves the lowest final objective while concentrating the search later in training. This suggests that BayesEvolve avoids unproductive archive collapse while still converting accumulated evidence into a focused search direction.

\section{Discussion}
\label{sec:discussion}

The experiments support three claims. First, explicit belief states improve sample efficiency relative to archive and heuristic-memory baselines. Second, the learned belief state is predictive on held-out candidates, rather than merely serving as a prompt artifact. Third, decision rules matter: most gains come from the posterior mean, while a decaying uncertainty bonus gives the best final performance by shifting from exploration to exploitation.

\paragraph{Limitations.}
Our current experiments use shifted BBOB-style numerical optimization tasks rather than full program or laboratory discovery. The GP belief state is predictive but imperfectly calibrated, and its scaling limits motivate sparse or neural surrogates for larger evaluation budgets. Finally, the diversity analysis shows productive concentration rather than sustained diversity; future work should study richer notions of semantic and structural diversity for program and scientific-hypothesis spaces.

\paragraph{Conclusion.}
We introduced BayesEvolve, a framework for autonomous discovery agents that transform experimental evidence into explicit predictive belief states. On shifted black-box optimization tasks, BayesEvolve improves over memory-guided LLM baselines, produces predictive belief estimates, and benefits from an annealed uncertainty-aware decision rule. These results support belief-state evolution as a useful principle for autonomous scientific discovery.

\bibliographystyle{unsrtnat}
\bibliography{bayesevolve}

\end{document}